\documentclass[final,5p,times,twocolumn]{elsarticle}

\usepackage{lineno,hyperref}
\usepackage{diagbox}
\usepackage{multirow}
\usepackage{amssymb}
\usepackage{hyperref}
\usepackage{dsfont}
\usepackage{bm}
\usepackage{amsmath}
\modulolinenumbers[5]

\journal{Journal of Image and Vision Computing}

\begin{document}

\begin{frontmatter}

\title{YinYang-Net: Complementing Face and Body Information for Wild Gender Recognition}

\author{Tiago Roxo\corref{mycorrespondingauthor}}
\ead{tiago.roxo@ubi.pt}
\author{Hugo~Proença}
\ead{hugomcp@di.ubi.pt}
\address{Department of  Computer Science, University of Beira Interior, Portugal}


\cortext[mycorrespondingauthor]{Corresponding author}


\begin{abstract}

Soft biometrics inference in surveillance scenarios is a topic of interest for various applications, particularly in security-related areas. However, soft biometric analysis is not extensively reported in \textit{wild conditions}. In particular, previous works on gender recognition report their results in face datasets, with relatively good image quality and frontal poses. Given the uncertainty of the availability of the facial region in wild conditions, we consider that these methods are not adequate for surveillance settings. To overcome these limitations, we: 1) present frontal and \textit{wild} face versions of three well-known surveillance datasets; and 2) propose YinYang-Net (YY-Net), a model that effectively and dynamically complements facial and body information, which makes it suitable for gender recognition in wild conditions. The frontal and \textit{wild} face datasets derive from widely used Pedestrian Attribute Recognition (PAR) sets (PETA, PA-100K, and RAP), using a pose-based approach to filter the frontal samples and facial regions. This approach retrieves the facial region of images with varying image/subject conditions, where the state-of-the-art face detectors often fail. YY-Net combines facial and body information through a learnable fusion matrix and a channel-attention sub-network, focusing on the most influential body parts according to the specific image/subject features. We compare it with five PAR methods, consistently obtaining state-of-the-art results on gender recognition, and reducing the prediction errors by up to 24\% in frontal samples. The announced PAR datasets versions and YY-Net serve as the basis for \textit{wild} soft biometrics classification and are available in \url{https://github.com/Tiago-Roxo}.
\end{abstract}


\end{frontmatter}


\section{Introduction}
%
%
%
%


Soft biometric cues are particularly useful in security-related issues and have increased in popularity in recent years. This is mainly motivated by their applicability in person identification, while requiring no cooperation from the subjects and being robust to low quality data \cite{jain2004soft, dantcheva2015else}. Furthermore, the availability of massive amounts of data acquired in uncontrolled conditions, provided by surveillance cameras and hand-held devices, has also contributed to enhancing the interest in this topic.

Gender is a particularly relevant soft label and its classification has been extensively reported in the literature, with various datasets announced~\cite{liu2015deep, rothe2018deep, cao2018vggface2, escalera2016chalearn,eidinger2014age}. However, we consider that gender recognition in \textit{wild conditions} is not yet extensively explored, with most of the previous methods frequently linking it to face analysis in relatively good quality data, without the challenges found in surveillance scenarios. Alternatively, gender information can be retrieved in video surveillance settings, using Pedestrian Attribute Recognition (PAR) datasets, with methods analyzing up to 51 attributes simultaneously and not the gender alone. This promotes the use of weighted losses for the different attributes and is an evident obstacle towards the optimization of results for each one.

Given that the face is intuitively a discriminative attribute to classify gender, our main goal is to develop an integrated framework that uses facial information, whenever possible, in surveillance conditions.
This is useful for scenarios where analysis is carried out mostly from frontal poses, such as airports, casinos, and border services. However, reliably retrieving the facial region from surveillance images can not be considered a solved problem. In this context, we start by creating \emph{frontal} versions of PAR datasets (PETA, PA-100K, and RAP) using pose data from Alphapose~\cite{fang2017rmpe, li2018crowdpose, xiu2018poseflow}. Then, we create \textit{wild} face datasets using a pose-based approach to define the head Region of Interest (ROI), from frontal PAR images. This approach is applicable in contexts with varying image quality and uncooperative subjects, outperforming state-of-the-art face detectors. Based on the varying image/subject qualities and face variability factors, the created \textit{wild} face datasets contrast with the state-of-the-art, showing its applicability for wild soft biometrics classification.

\begin{figure}[!tb]
\centering
\includegraphics[scale=0.8]{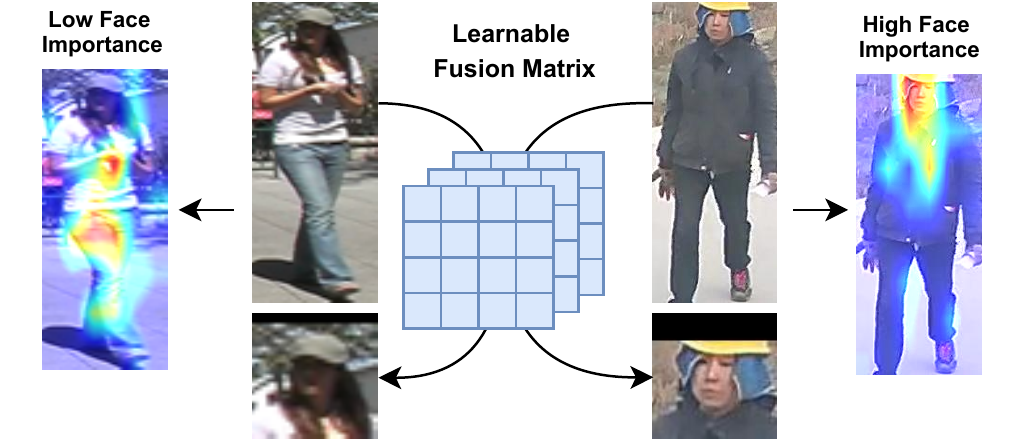}
\caption{Illustration of the effect of the learnable fusion matrix, a major component of YY-Net, as an attention mechanism with respect to the (non)availability of facial information.}
\label{fig:fusion-matrix}
\end{figure}

To effectively classify gender, using the created datasets, we propose YinYang-Net (YY-Net), a model that combines face and body-based features through a learnable fusion matrix, promoting weighted conjunction of both types of information. Fig.~\ref{fig:fusion-matrix} displays the effect of that matrix, with respect to the availability of face information. Moreover, we apply a residual link with body features, using a channel attention sub-network, to promote a weighted face-body combination. This contributes to better focus on the most influential body parts for gender recognition. 

According to our experiments, the proposed model improves gender recognition results in all frontal PAR datasets, when compared to five state-of-the-art PAR methods. Furthermore, in non-frontal images, where the facial region is unavailable, YY-Net uses body information and not the weighted combination of face-body features. In this context, it also outperforms the baseline PAR methods, in all datasets. 

In summary, the major contributions of this paper are:
\begin{itemize}
    \item We present frontal and \textit{wild} face versions of well-known PAR datasets (PETA, PA-100K, and RAP), released as meta-data, yielding from a pose processing approach. Given their face variability factors, these \textit{wild} face datasets contrast with the state-of-the-art, representing more challenging sets to soft biometrics applications;
    \item We propose YY-Net, an attention-based model for \textit{wild} gender recognition, complementing face and body information, that surpasses the state-of-the-art in PAR datasets, particularly in frontal samples.\\
\end{itemize}

The remainder of this paper is organized as follows: Section \ref{sec:related-work} summarizes the most relevant gender recognition works; Section \ref{sec:proposed-method} describes the proposed model; Section \ref{sec:proposed-dataset} describes the proposed datasets, and Section \ref{sec:experiments} discusses the results obtained. The main conclusions and future work are presented in Section \ref{sec:conclusion}.

\begin{figure*}[!tb]
\centering
\includegraphics[width=0.95\textwidth]{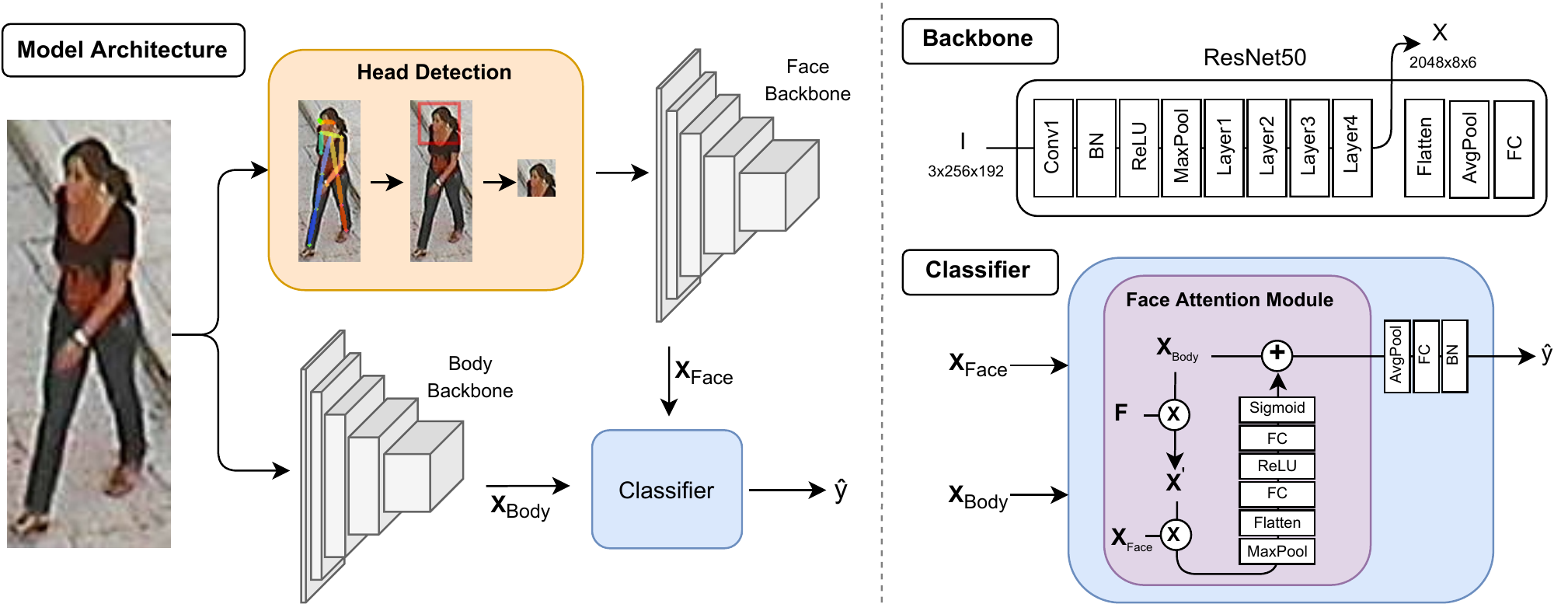}
\caption{
Cohesive perspective of the YY-Net, divided into three parts: \textit{1) Model Architecture}; \textit{2) Backbone}; and \textit{3) Classifier}. The \textit{Model Architecture} displays our processing flow, where the input data feeds two different backbones: 1) for the body; and 2) for facial information. The facial regions are obtained according our head detection approach, using pose information. The backbones derive from ResNet50, with classifier-related layers ignored, as shown in the \textit{Backbone} portion. The combination of the face and body inputs is done through a classifier, which uses the proposed Face Attention Module to fuse both types of information effectively.
}
\label{fig:main_image}
\end{figure*}

\section{Related Work}
\label{sec:related-work}

\subsection{Face-Based Gender Recognition}

The face is one of the most discriminative aspects to classify gender, which culminated in several datasets designed for this task~\cite{liu2015deep, rothe2018deep, cao2018vggface2, escalera2016chalearn}. Given the nature of gender and age estimation, simple Convolutional Neural Network (CNN) models (with only five layers) can be used for gender recognition~\cite{levi2015age}. However, more complex architectures have recently been used, such as those based on Residual Networks of Residual Networks (RoR)~\cite{zhang2017age}, which have outperformed other CNN architectures.

Some works have focused on complementing the face images with facial attributes analysis. 
Han \textit{et al.}~\cite{han2017heterogeneous} proposed a Deep Multi-Task Learning (DMTL) framework for joint estimation of face attributes, considering their correlation and heterogeneity for various demographic attributes (gender, age, race).  Ranjan \emph{et al.} \cite{ranjan2017hyperface} developed HyperFace, a multi-task framework based on Alexnet \cite{krizhevsky2012imagenet} for face detection, fused with a CNN for face landmarks detection and gender recognition. Liu \emph{et al.} \cite{liu2015deep} combined a Support Vector Machine (SVM), for face attribute classification, with a CNN for appropriately resized face localization and another for distinguishing face identities.

The complement of particular face attributes for gender classification is also a topic of research. 
Datcheva and Br{\'e}mond~\cite{dantcheva2016gender} analyzed smile-face dynamics in gender estimation, relating its importance with the subject's age and demonstrating the complement of dynamic and appearance-based features. Lapuschkin \textit{et al.}~\cite{lapuschkin2017understanding} examined which facial attributes actually impact age and gender prediction, using a Layer-wise Relevance Propagation. In the same topic,  Yaman \textit{et al.}~\cite{yaman2019multimodal} used ear appearance to aid gender estimation in profile face images, employing domain adaptation.  Ryu \emph{et al.} presented InclusiveFaceNet \cite{ryu2017inclusivefacenet}, proposing the inclusion of race and gender for face attribute detection.

\subsection{Body-Based Gender Recognition}

Regarding wild gender recognition, where the focus is on the whole body and not on the facial region alone, we present works using PAR datasets~\cite{deng2014pedestrian,liu2017hydraplus,li2016richly}, a context that more closely relates to surveillance scenarios. Some works opt to analyze the relation of different body parts: Wang \emph{et al.} \cite{wang2017attribute} proposed a model based on a Recurrent Neural Network (RNN) encoder-decoder framework, using Long Short-Term Memory (LSTM) as a recurrent neuron, with joint recurrent learning via images division into six horizontal strips; Zhao \emph{et al.} \cite{zhao2018grouping} also used a similar partition strategy, combining LSTM and BN-Inception \cite{ioffe2015batch} networks.

The importance of various attributes is also reported in the literature, namely on the relation of attributes towards pedestrian attribute recognition.  Learning the relationship, dependency, and correlation between attributes and using this information to accurately predict attributes have been the focus of some works~\cite{li2015multi,lin2019improving}. Similarly, using a hierarchical feature embedding framework~\cite{yang2020hierarchical} has proven to promote finer-grained clustering, translating into increased attribute recognition accuracy. However, enhancing the localization of attribute-specific areas, typically adopted by state-of-the-art methods, may not necessarily improve performance~\cite{jia2020rethinking}. Image and subject-based features are also aspects to consider for accuracy improvement, given that their importance in gender classification varies with image quality and face availability~\cite{roxo2021gender}.

Pose information and body part localization are also influential for pedestrian attribute recognition. Li \textit{et al.}~\cite{li2018pose} used pose data to aid in attribute body part localization, fusing features at multiple levels for attribute recognition. Zhang \textit{et al.}~\cite{zhang2020deep} proposed a Deep Template Matching method to capture body parts features, complemented by pose keypoints to guide discriminative cues learning. Liu \textit{et al.}~\cite{liu2018localization} used body ROI and assigned attribute-specific weights based on extracted ROI proposals and attribute localization. Yaghoubi \textit{et al.}~\cite{yaghoubi2019region} used body keypoint estimation to define attribute region of interest, while also evaluating pose effect.

\subsection{Attention-Based Gender Recognition}

Various works base their approaches on attention mechanisms. Effective attention maps at different scales~\cite{sarafianos2018deep}, analyzing relevant image patches~\cite{rodriguez2017age}, or focusing on different feature levels~\cite{tang2019improving} are all reported methodologies to discover discriminative regions for attribute classification. Alternatively, using an attention consistency loss~\cite{guo2019visual} based on heatmaps difference of original and flipped images can yield good performances.

Attribute relation was also the basis of attention approaches~\cite{li2020attention,fan2020correlation,zhao2019recurrent,ji2020pedestrian,gao2019pedestrian}. Li \textit{et al.}~\cite{li2020attention} used Convolutional LSTM (ConvLSTM) for spatial and semantic correlation between attributes, combining it with channel attention to adjust the weight of relevant channel features adaptively. Zhao \textit{et al.}~\cite{zhao2019recurrent} proposed the use of a Recurrent Attention Model (also based on ConvLSTM), mining the correlations among different spatially related attributes. Ji \textit{et al.}~\cite{ji2020pedestrian} announced an encoder-decoder with multiple LSTM, aiming to exploit more contextual knowledge, mining deeper relations between images and attributes. Gao \textit{et al.}~\cite{gao2019pedestrian} proposed a two-stage training, complementing semantic segmentation with attribute recognition, relating attributes in an attention-based manner between both stages.

\section{YinYang-Net}
\label{sec:proposed-method}

An overview of YY-Net is shown in Fig.~\ref{fig:main_image}. The input is processed in two ways: 1) retrieving body information; and 2) retrieving face information. We obtain facial information by first processing the image to obtain pose data and then using a head ROI detection approach, described in Section~\ref{sec:head-detect}. We use Alphapose~\cite{fang2017rmpe, li2018crowdpose, xiu2018poseflow} to collect pose information. Both body and face information are retrieved from a modified ResNet50~\cite{he2016deep}, via feature maps at \textit{Layer 4}.  Finally, we pass both feature maps to a classifier for gender recognition.

\subsection{Network Architecture}
\label{network-architecture}

The key concept of our approach is to combine face and body information to classify gender. The facial region is intuitively a discriminative attribute for gender recognition, and its importance is linked to the subject pose. To achieve our goal, we use a ResNet50, pretrained in ImageNet~\cite{deng2009imagenet}, for feature extraction of both face and body images. To better control features combination, we retrieve features from a modified ResNet50 model, with \textit{Flatten}, \textit{Pooling}, and \textit{FC} layers removed. 
We refer the modified ResNet50 as \textit{backbone}. We use two different backbones for face and body image processing to fine-tune extraction based on the body region inputted.

Given an input image $\bm{I}$, we denote the features extracted as $\phi_{i} (\bm{I}) \in \mathbb{R}^{H_{i}\times W_{i}\times C_{i}},~~i \in \{face, body\} $. For a $3\times256\times192$ input image, we obtain a $H_{i}\times W_{i}\times C_{i}$ of $2048\times8\times6$, for both values of $i$. We represent our features as $\bm{X}$:

\begin{equation}
\bm{X}_{i} = \phi_{i} (\bm{I}),~~i \in \{face, body\},
\end{equation}

where $\phi_{i}$ is the backbone for a given input type. Given that our framework aims to use facial information effectively, Fig.~\ref{fig:main_image} displays the steps involved for gender recognition in frontal images. However, we are able to adapt our framework to consider any pose images through pose data processing, following our head detection approach. Representing $\theta$ as said approach, we define gender recognition of the $n^{th}$ input as:

\begin{equation}
  \hat{y}_{n} =
    \begin{cases}
      f (\rho(\bm{X}_{face_{n}}, \bm{X}_{body_{n}})), & \text{if $\mathds{1}_{\theta(\bm{I}_n) = Frontal}$}  \\
      f (\bm{X}_{body_{n}}), & \text{ otherwise}, \\
    \end{cases}       
\end{equation}

where $f$ is a sequence of \textit{Average Pooling}, \textit{FC}, and \textit{Batch Normalization} layers, $\rho$ is our Face Attention Module (FAM), $\mathds{1}$ is the indicator function, and $\bm{X}$ are the  features extracted from a given body part.

\subsection{Face Attention Module}
\label{subsec:fam}

Having the body and face features extracted, we combine them efficiently by promoting the focus of YY-Net in the most relevant regions of the body. We design a weighted face-body feature combination by using a learnable weights matrix: $\textbf{F} \in \mathbb{R}^{H_{i}\times W_{i}\times C_{i}}$. We denote the features combination $\bm{X}_{c}$  as:

\begin{equation}
\bm{X}_{c} = \bm{X}_{body} \odot \bm{F} \odot \bm{X}_{face},
\label{eq:x_combined}
\end{equation}

where $\odot$ represents the Hadamard product, $\bm{F}$ is our fusion matrix and $\bm{X}$ are the extracted features. The importance of each body part is expected to vary depending on the image quality, person pose, and partial occlusions. For this reason, we add a channel-attention sub-network, based on the Squeeze-and-Excitation block~\cite{hu2018squeeze}, to modulate inter-channel dependencies. The combined input $\bm{X}_{c}$ passes through a series of linear and nonlinear layers, as shown in the \textit{Classifier} part of Fig.~\ref{fig:main_image}, yielding a weight vector for the feature importance across channels. Then, we induce a residual link of this vector with body features to promote information complementary. We express the FAM output as:

\begin{equation}
\rho = \psi (\bm{X}_{c}) + \bm{X}_{body},
\end{equation}

where $\psi$ is the sequence of linear and nonlinear layers previously described, $\bm{X}_{c}$ refers to the result of (\ref{eq:x_combined}), and $\bm{X}_{body}$ corresponds to the extracted body features.

\subsection{Implementation Details}
Images are resized to $256\times192$ with random horizontal mirroring as inputs. The Stochastic Gradient Descent (SGD) algorithm is used for training, with momentum of $0.9$ and weight decay of $0.0005$.  The initial learning rate is set to $0.01$ for the backbones and classifier. Plateau learning rate scheduler is used with a reduction factor of $0.1$ and a patience epoch number of $4$. Batch size is set to $32$ and the number of training epochs is set to $30$, using Binary Cross Entropy with Logits (BCELogits) as loss function.

\section{Frontal PAR and Face Datasets}
\label{sec:proposed-dataset}

\begin{figure}[!tb]
\centering
\includegraphics[scale=0.5]{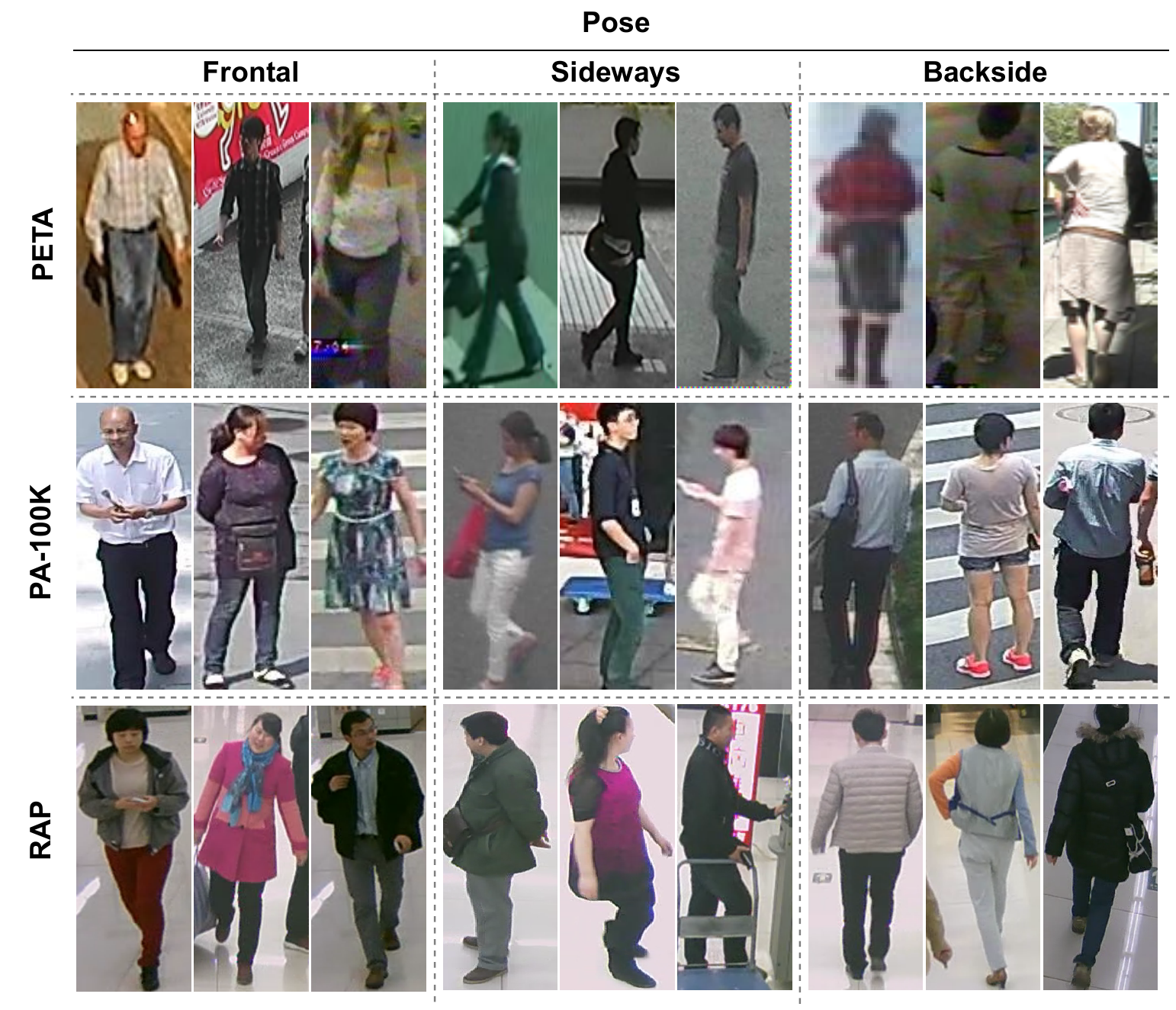}
\caption{Examples of images considered as \textit{frontal}, \textit{sideways} and \textit{backside}. Each row is related to one of the three PAR datasets and each column regards examples of the respective pose.}
\label{fig:people_orientation}
\end{figure}

We create frontal and \textit{wild} face versions of three well-known PAR datasets. The frontal images are obtained obtained from a division of three possible values: \textit{frontal}, \textit{sideways}, and \textit{backside}. We differentiate these values based on pose information from Alphapose~\cite{fang2017rmpe, li2018crowdpose, xiu2018poseflow}: if the rightmost shoulder, from the top left of the image, is the left shoulder, the subject faces forward (\textit{frontal}); otherwise, the subject is facing backward (\textit{backside}). Regarding \emph{sideways} classification, we attribute this label to subjects with a shoulder length, with respect to the upper body height, less than 0.5. Examples of the pose classes considered are presented in Fig. \ref{fig:people_orientation}. 

\subsection{Head Detection}
\label{sec:head-detect}

Head ROI detection also uses pose information, cropping head regions from frontal images. Initially, we obtain the coordinates $x$ and $y$ of the right and left ears, considering the head's central point as the arithmetic mean of such coordinates. Then, head ROIs are drawn using the top-left and bottom-right bounding boxes coordinates, centered in the head's central point, with the head ROI height as $\frac{2}{9}$ of the whole body silhouette height. Head bounding boxes drawing is based on Detectron~\cite{detectron2}.

\begin{figure}[!tb]
\centering
\includegraphics[scale=0.5]{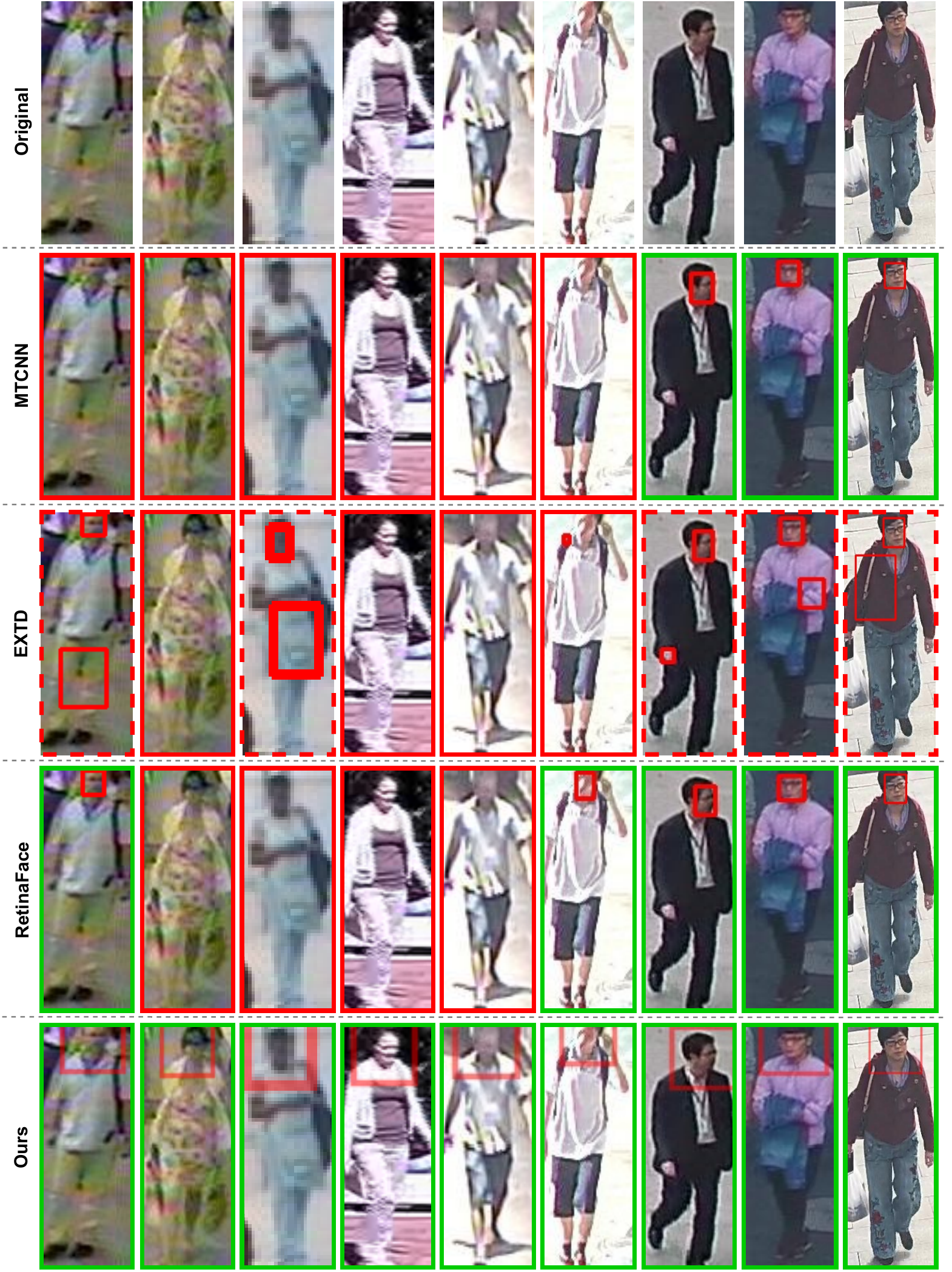}
\caption{Comparison of the head ROI detection results between MTCNN, EXTD, RetinaFace, and our approach, in images with varying quality, head pose and blurriness. Green boxes denote cases where the method accurately locates the face, and red boxes denote the opposite. Dotted red boxes refer to cases where the model over-predicted the number of existent faces. All images are resized to the same resolution for visualization purposes.}
\label{fig:head_detection_comparison}
\end{figure}

This head ROI detection approach can retrieve head regions from frontal images with varying image quality and/or partial occlusions.  Other face detectors, such as MTCNN~\cite{zhang2016joint}, EXTD~\cite{yoo2019extd}, and RetinaFace~\cite{deng2020retinaface}, fail to detect head region as reliably as the used approach in the considered (wild) image conditions. To show the versatility of this method, Fig.~\ref{fig:head_detection_comparison} presents some examples of head ROI detection in frontal PAR images. In our head detection comparison, we use the default implementation of all methods.

\begin{figure*}[!tb]
\centering
\includegraphics[scale=0.8]{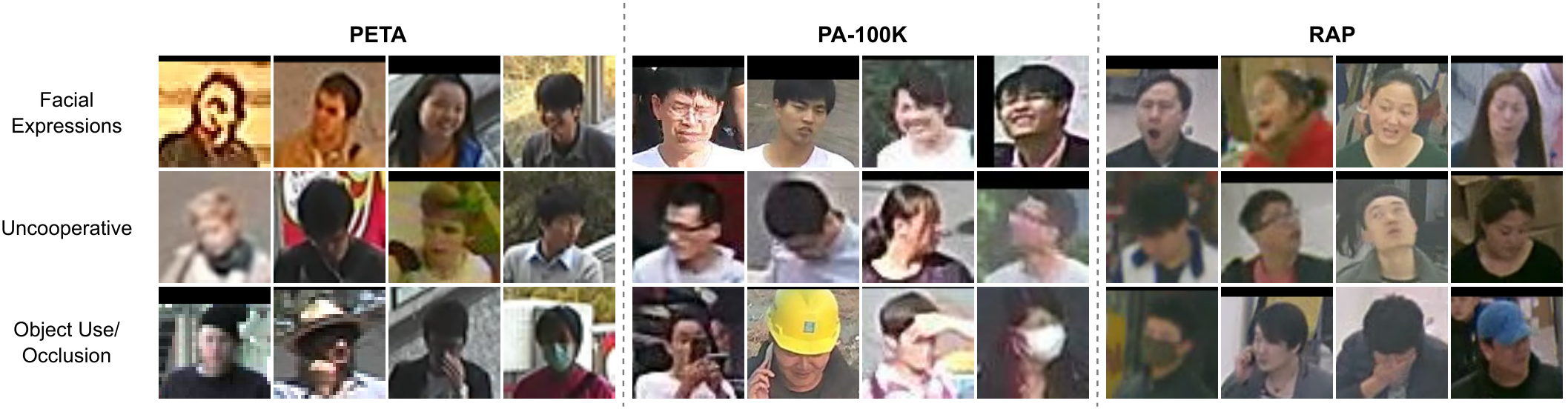}
\caption{Examples of image quality variability in \textit{wild} face datasets. The datasets contain subjects with varying facial expressions, levels of cooperativeness, and partial occlusions. Each row refers to one variability factor, and each group of four columns regards examples of a PAR dataset.}
\label{fig:face_dataset}
\end{figure*}

We observe that MTCNN only detects faces in higher quality images, failing to detect in lower quality ones. This is expected since MTCNN was not designed to target ``small faces" specifically, unlike EXTD and RetinaFace. Nonetheless, these methods fail to detect faces in low quality images reliably.  EXTD was able to detect faces in more images but outputs more incorrect detections as well. RetinaFace accurately detects faces in more images than MTCNN but is not able to detect faces when quality is subpar, with a background color similar to the person (second and fourth column) or with blurry faces (third and fifth column). Our approach detects the facial region in all presented scenarios, regardless of the image quality. This is an important aspect, given that we intend to provide our model with facial images from surveillance scenarios, with varying image quality. Furthermore, we use the described approach to create \textit{wild} face datasets, representing a challenging set for soft biometrics classification. The variability encountered in these datasets is displayed in Fig.~\ref{fig:face_dataset}.

\subsection{Data Analysis}

To prove that frontal datasets preserve the main characteristics of the original one, we analyze the major image-based factors that influence gender recognition. We consider \textit{resolution}, \textit{luminosity}, and \textit{blurriness} as the main image-based features: \textit{resolution} is retrieved from image width and height multiplication; for \textit{luminosity}, we use the red, green, and blue value channels to measure the perceived brightness~\cite{lumino}; and \textit{blurriness} yields from the convolution of the images with a Laplacian kernel, taking the variance as result. The image-based feature analysis of the used PAR datasets, and respective frontal versions, is shown in Table~\ref{datasets-comparison}.

The obtained results suggest that frontal datasets preserve the characteristics of the original version. As such, they are not easier for gender recognition, and the accuracy variance of models on these datasets (when compared to the original ones) derives from pose and face access. Furthermore, given the resolution and blurriness values, RAP can be considered the dataset with the highest quality, while PETA is the lowest quality one.

\begin{table}[!tb]
    \centering
    \footnotesize
    \renewcommand{\arraystretch}{1.05}
    \caption{Dataset image-based feature analysis. Values are normalized for the combination of all dataset values, for each feature. Datasets named with \textit{Frontal} suffix refer to datasets composed solely of frontal images.}
    \begin{tabular}{|c|*{5}{c|}}\hline
        \textbf{Dataset} & 
        \textbf{Resolution} &
        \textbf{Luminosity} &
        \textbf{Blurriness} 
        \\\hline\hline
        
        PETA &
        0.037 $\pm$ 0.024 &
        0.432 $\pm$ 0.100 &
        0.120 $\pm$ 0.148 \\
        
        PETA\textsubscript{Frontal} &
        0.037 $\pm$ 0.023 &
        0.437 $\pm$ 0.101 &
        0.140 $\pm$ 0.158 \\
        
        \hline
        PA-100K & 
        0.062 $\pm$ 0.063 &
        0.449 $\pm$ 0.126 & 
        0.095 $\pm$ 0.089 \\

        PA-100K\textsubscript{Frontal} & 
        0.061 $\pm$ 0.064 &
        0.459 $\pm$ 0.127 & 
        0.100 $\pm$ 0.090 \\
        
        \hline
        RAP & 
        0.131 $\pm$  0.083 &
        0.407 $\pm$  0.107 & 
        0.022 $\pm$  0.013 \\
        
        RAP\textsubscript{Frontal} & 
        0.138 $\pm$  0.090 &
        0.416 $\pm$  0.108 & 
        0.023 $\pm$  0.013 \\
        \hline
        
    \end{tabular}
    \label{datasets-comparison}
\end{table}

Another conducted experiment was the evaluation of image number ratio between \textit{test} and \textit{all} (train + test) images.
Furthermore, we assess the image number ratio of frontal and PAR datasets, for the test and train set, to verify if the pose division was balanced. We present the ratios analyzed in Table~\ref{table:all-frontal-dataset-comparison}. The test/all images ratios are very similar between all PAR datasets and their frontal versions. Additionally, frontal/PAR ratios are similar between datasets and around $\frac{1}{3}$, which is expected given that the dataset pose division is based on three values (\textit{frontal}, \textit{sideways}, and \textit{backside}).

Regarding the \textit{wild} face datasets, all the ratios presented for frontal datasets are analogous for the face ones. The resolution, luminosity, and blurriness of face datasets are similar to the frontal dataset values, given that the facial regions were retrieved from frontal images.

\begin{table}[!tb]
    \centering
    \footnotesize
    \renewcommand{\arraystretch}{1.05}
    \caption{Number of images, test/all image number ratio for each dataset, and frontal/PAR image number ratio, for train and test set. Frontal datasets are represented with the subscript term \textit{Frontal} and test/all ratios are represented by Dataset Ratio.}
    \begin{tabular}{|c|*{5}{c|}}\hline
        \multirow{2}{*}{\textbf{Dataset}} & 
        \multicolumn{2}{c|}{\textbf{Train}} &
        \multicolumn{2}{c|}{\textbf{Test}} & 
        \makebox[3em]{\textbf{Dataset}} \\
        &
        \makebox[3em]{Images} & 
        \makebox[3em]{Ratio} &
        \makebox[3em]{Images} & 
        \makebox[3em]{Ratio} &
        \makebox[3em]{\textbf{Ratio}}
        \\\hline\hline
        
        PETA & 11400 & \multirow{2}{*}{0.379} & 7600 & \multirow{2}{*}{0.383} & 0.400 \\
        PETA\textsubscript{Frontal} & 4318 & & 2918 & & 0.403 \\
        
        \hline
        PA-100K & 90000 & \multirow{2}{*}{0.379} & 10000  & \multirow{2}{*}{0.354} & 0.100 \\
        PA-100K\textsubscript{Frontal} & 34128 & & 3539 & & 0.093 \\
        
        \hline
        RAP & 33268 & \multirow{2}{*}{0.339} & 8317  & \multirow{2}{*}{0.346} & 0.200 \\
        RAP\textsubscript{Frontal} & 11275 & & 2880 & & 0.203 \\
        \hline
        
    \end{tabular}
    \label{table:all-frontal-dataset-comparison}
\end{table}

\section{Experiments}
\label{sec:experiments}

\subsection{Datasets, Methods, and Evaluation Metrics}

Our experiments are carried out in three state-of-the-art PAR datasets: PETA \cite{deng2014pedestrian}, PA-100K \cite{liu2017hydraplus}, and RAP \cite{li2016richly}.  The PETA dataset has 19,000 images, divided into three subsets: 9,500 for training, 1,900 for validation, and 7,600 for testing. The PA-100K dataset is composed of 100,000 images from outdoor surveillance cameras. It is split into 80,000 images for training, 10,000 for validation, and 10,000 for testing. The RAP dataset contains 41,585 images. Following the official protocol \cite{li2016richly}, we split the dataset into 33,268 training images and 8,317 test images. Furthermore, we carried out experiments in the proposed frontal datasets, consisting of frontal images of the described PAR datasets.

The PAR methods used in our experiments are VAC \cite{guo2019visual}, APR \cite{lin2019improving}, ALM \cite{tang2019improving}, StrongBase \cite{jia2020rethinking}, and DeepMar \cite{li2015multi}. All models are evaluated using the original implementations with minor adaptations to only classify gender and not all pedestrian attributes.

To present our results, we use the label-based metric mean Accuracy ($mA$)~\cite{li2016richly}, which is the mean classification accuracy of the positive and negative samples for each attribute, averaged over all attributes:

\begin{equation}
mA = \dfrac{1}{2n} \sum_{i=1}^{m} \left(\dfrac{tp_i}{p_i} + \dfrac{tn_i}{n_i} \right),
\end{equation}

where \textit{n} is the number of examples, \textit{m} the number of attributes, \textit{$p_i$} is the number of positive examples and \textit{$tp_i$} is the number of correctly predicted positive examples for the $i^{th}$ attribute; \textit{$tn_i$} and \textit{$n_i$} are defined analogously. We adapt this metric to only consider the gender attribute, from the PAR datasets. 

\subsection{Gender Recognition}

Given that our focus is to improve wild gender recognition, combining facial and body information, we start by assessing the performance of YY-Net in frontal PAR datasets. We compare it with five state-of-the-art PAR methods, presenting our results in Table \ref{datasets-frontal-methods}.

\begin{table}[!tb]
    \centering
    \footnotesize
    \renewcommand{\arraystretch}{1.05}
    \caption{Gender $mA$ of different models on frontal PAR datasets. The outperforming method for each dataset is shown in bold.}
    \begin{tabular}{|c|*{5}{c|}}\hline
        \textbf{Methods} & 
        \makebox[5.5em]{\textbf{PETA\textsubscript{Frontal}}} &
        \makebox[5.5em]{\textbf{PA-100K\textsubscript{Frontal}}} &
        \makebox[5.5em]{\textbf{RAP\textsubscript{Frontal}}} 
        \\\hline\hline
        
        ALM~\cite{tang2019improving} & 
        91.27 &
        91.23 &
        95.71 \\
        
        APR~\cite{lin2019improving} & 
        92.49 &
        91.63 &
        95.80 \\
        
        DeepMar~\cite{li2015multi} & 
        91.09 &
        90.75 &
        95.96 \\
        
        VAC~\cite{guo2019visual} & 
        91.17 &
        90.15 &
        94.49 \\
        
        StrongBase~\cite{jia2020rethinking} & 
        92.62 &
        92.05 &
        96.14 \\
        \hline
        
        \textbf{YY-Net} & 
        \textbf{93.45} & 
        \textbf{92.79} &
        \textbf{97.07}  \\
        \hline
        
    \end{tabular}
    \label{datasets-frontal-methods}
\end{table}

Our approach is better than all considered methods, which corroborates the importance of effectively combining face and body images in gender classification. When comparing with the best result for each dataset, we improve gender accuracy by 0.83\%, 0.74\%, and  0.93\% in PETA, PA-100K, and RAP, respectively. The improvement in all datasets demonstrates the versatility of YY-Net, corresponding to an error reduction of 11.25\%, 9.31\%, and 24.09\%, respectively. The increased error reduction displayed in RAP could be linked to its higher image quality, relative to the other PAR datasets. High quality images might benefit more from effective facial and body information conjunction, giving the increased body details, culminating in improved relevant body portion focus.

With the intent to analyze the adaptability of our model to deal with unconstrained pose scenarios, we evaluate its performance in PAR datasets. We present, in Table~\ref{datasets-methods}, the accuracy of the analyzed state-of-the-art PAR methods, for each dataset. 

\begin{table}[!tb]
    \centering
    \footnotesize
    \renewcommand{\arraystretch}{1.05}
    \caption{Gender $mA$ of different models on PAR datasets. The outperforming method for each dataset is shown in bold.}
    \begin{tabular}{|c|*{5}{c|}}\hline
        \textbf{Methods} & 
        \makebox[3.5em]{\textbf{PETA}} &
        \makebox[3.5em]{\textbf{PA-100K}} &
        \makebox[3.5em]{\textbf{RAP}} 
        \\\hline\hline
        
        ALM~\cite{tang2019improving} & 
        92.28 &
        90.34 &
        95.69 \\
        
        APR~\cite{lin2019improving} & 
        92.84 &
        90.05 &
        96.20 \\
        
        DeepMar~\cite{li2015multi} & 
        92.33 &
        90.39 &
        96.40 \\
        
        VAC~\cite{guo2019visual} & 
        92.85 &
        91.05 &
        96.59 \\
        
        StrongBase~\cite{jia2020rethinking} & 
        93.13 &
        90.77 &
        96.74 \\
        \hline
        
        \textbf{YY-Net} & 
        \textbf{93.39} & 
        \textbf{91.20} &
        \textbf{96.86}  \\
        \hline
        
    \end{tabular}
    \label{datasets-methods}
\end{table}

Our results show that the combination of face and body information, even in the context of unconstrained poses, contributes to slight improvements. This effect was observed for all three datasets.  Compared to the second best method, YY-Net improved by 0.26\%, 0.15\%, and 0.12\% in PETA, PA-100K, and RAP, respectively, which corresponds to an error reduction of 3.79\%, 1.68\%, and 3.68\%. The results suggest that YY-Net is reliable in frontal pose images and applicable in wilder contexts, with varying poses. 

\subsection{Within and Cross-Domain Performance}

To assess if the performance improvements in PAR datasets derive essentially from more accurate results for frontal images, we evaluate YY-Nets in frontal datasets, using the weights from PAR training and compare it to YY-Nets trained solely on frontal images. We report our results in Table~\ref{datasets-frontal-methods-train-par}, exploring \textit{within} and \textit{cross-domain} settings.

\begin{table}[!tb]
    \centering
    \footnotesize
    \renewcommand{\arraystretch}{1.05}
    \caption{Gender $mA$ of YY-Net in frontal datasets, when trained in PAR datasets and in its frontal version. Frontal datasets are represented with the subscript term \textit{Frontal}. The highest evaluation accuracy for each dataset is shown in bold.}
    \begin{tabular}{|c|*{5}{c|}}\hline
        \backslashbox{Train}{Test} & 
        \makebox[5em]{PETA\textsubscript{Frontal}} &
        \makebox[5em]{PA-100K\textsubscript{Frontal}} &
        \makebox[5em]{RAP\textsubscript{Frontal}} 
        \\\hline\hline
        
        PETA & 
        \textbf{93.88} & 
        78.67 &
        81.47  \\
        
        PETA\textsubscript{Frontal} & 
        93.45 & 
        77.38 &
        78.09  \\
        
        \hline
        PA-100K & 
        79.02 & 
        92.70 &
        89.51  \\
        
        PA-100K\textsubscript{Frontal} & 
        77.98 & 
        \textbf{92.79} &
        89.10  \\
        
        \hline
        RAP & 
        73.06 & 
        75.16 &
        96.95 \\
        
        RAP\textsubscript{Frontal} & 
        73.51 & 
        77.94 &
        \textbf{97.07} \\
        \hline
        
    \end{tabular}
    \label{datasets-frontal-methods-train-par}
\end{table}

This experiment shows that models trained on different datasets have distinct performances, linked to dataset length and quality differences. If we analyze the \textit{within-domain}, only PETA has a significant difference between PAR and frontal training models, with the PAR model obtaining better performance. Models trained in PAR datasets do not interact with more frontal images than those trained in frontal datasets. However, they are trained in more images and with more varying body poses. This contributes to better body information processing which, in conjunction with our face combination approach, translates into a more accurate gender recognition. In this scenario, PETA training is more beneficial than only PETA\textsubscript{Frontal}, given that it is the dataset with the lowest number of training images. Regarding the RAP and PA-100K, their frontal versions contain enough training images that PAR training does not translate into increased \textit{within-domain} performance.

When we analyze the \textit{cross-domain} performance, we observe that PAR training is more beneficial to PETA and PA-100K, with RAP having better performance with frontal training only. These differences could be associated with image quality, where RAP is the highest quality dataset, given our results in Table~\ref{datasets-comparison}. This is particularly relevant, given that high-quality images have increased face importance, particularly in frontal images. As such, training solely in frontal images may promote a more effective face data processing, translating into a better face-body combination. In this case, training for various poses might dissipate this aspect leading to a poorer generalization, expressed by worse \textit{cross-domain} performance in RAP. For PETA and PA-100K, more training data (PAR) promotes better generalization, as the \textit{cross-domain} results demonstrate.

\subsection{Face Attention Influence}

\begin{figure}[!tb]
\centering
\includegraphics[scale=0.55]{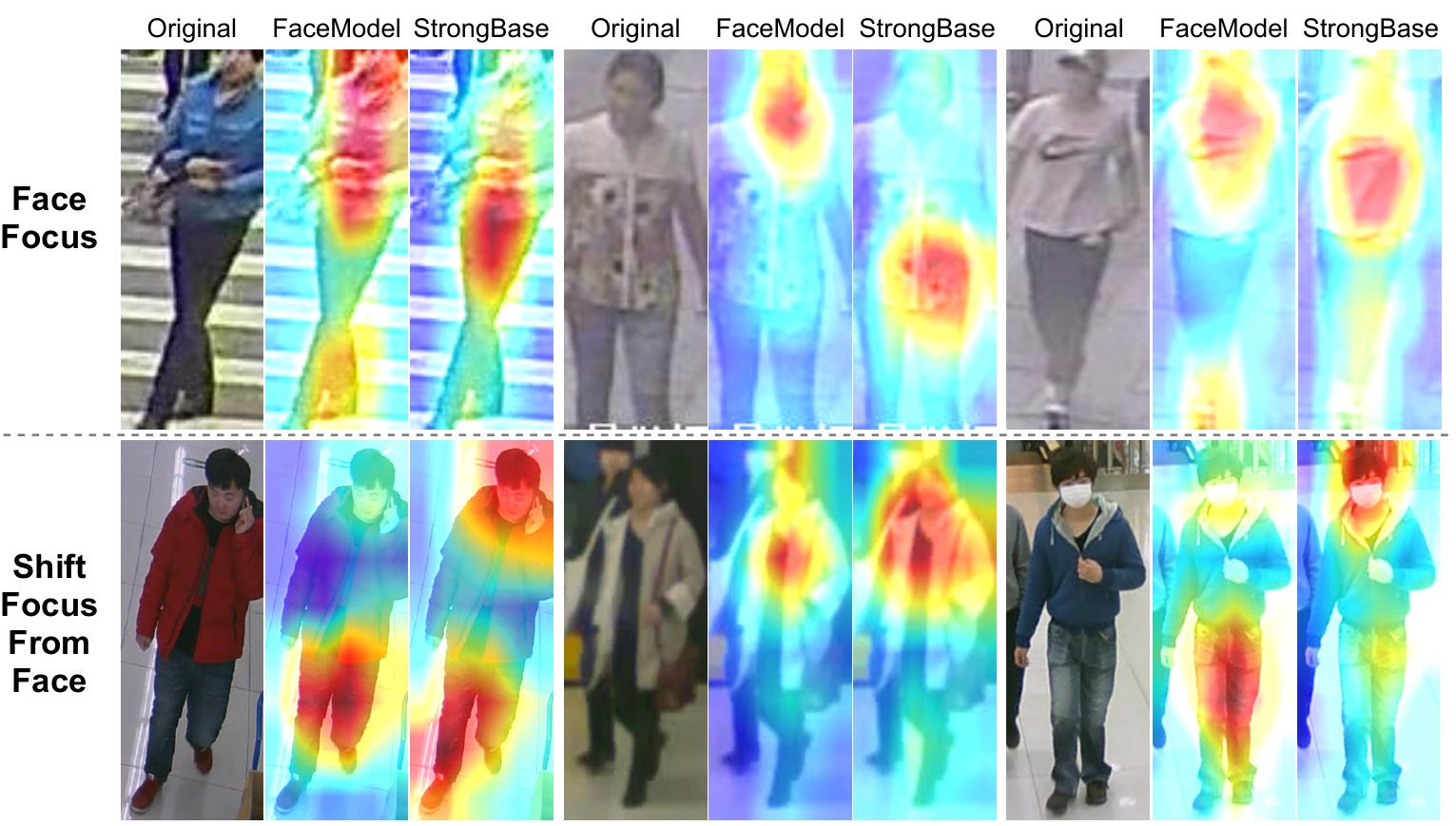}
\caption{Comparison of StrongBase and YY-Net attention focus from images where YY-Net correctly classifies gender and StrongBase does not. Each row is related to one of the categories of diverging focus.}
\label{fig:correct_front_fail_strong}
\end{figure}

Given that YY-Net uses face information to aid in gender recognition, we assess whether this approach would, implicitly, focus more on faces to perform its task. To achieve our goal, we compare Class Activation Maps (CAM)~\cite{zhou2016learning} from our model with the StrongBase's ones, from images where StrongBase exclusively misclassifies. This is the chosen method since it outperformed the remaining ones. We group all three datasets and present the CAM of representative images in Fig.~\ref{fig:correct_front_fail_strong}.

\begin{figure}[!tb]
\centering
\includegraphics[scale=0.55]{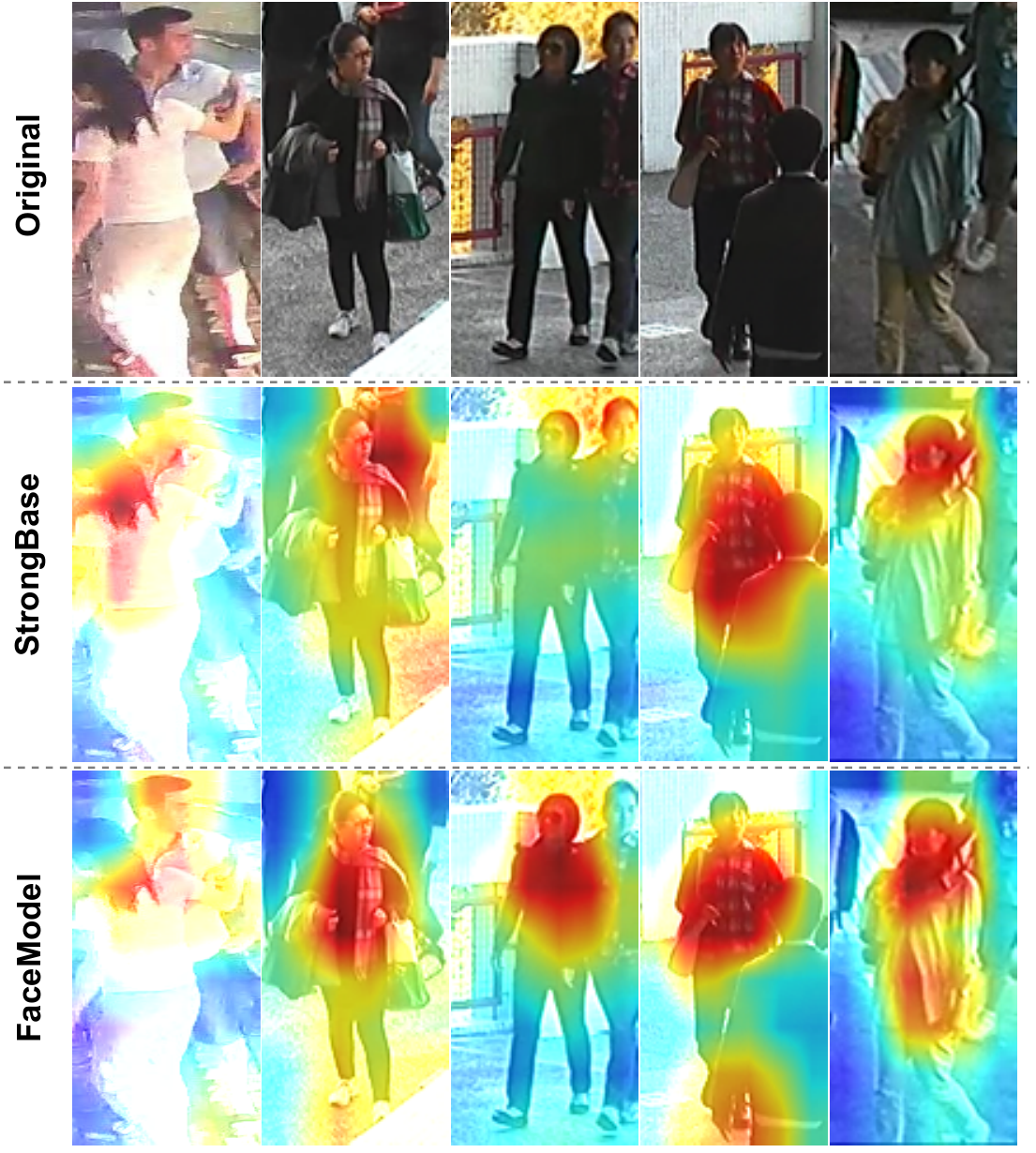}
\caption{Crowd effect on attention focus from StrongBase and YY-Net. Each column displays an example where YY-Net accurately classifies gender and StrongBase misclassifies it.}
\label{fig:crowd_compare_strong_front}
\end{figure}

The displayed examples suggest that our FAM does not force a predominant face focus but does contribute to a more coherent face-body combination. In the first row (\textit{Face Focus}), we observe a hip/upper body focus from StrongBase. In these cases, the facial region is a more discriminative factor (relative to the hip/upper body region), which is the focus of YY-Net. However, there are instances where the facial region may difficult/misguide gender recognition. In the second row, we observe cases where the facial region is not easily disclosed or presents unisex qualities. In these cases, analyzing other body portions may be beneficial, which is the approach taken by our model. StrongBase opts to focus more on the neck/facial region, contributing to gender misclassification. Another scenario where YY-Net has better performance is in crowded scenarios. We display images with multiple subjects where our model accurately classifies gender and StrongBase does not, in Fig.~\ref{fig:crowd_compare_strong_front}.

Influenced by FAM, YY-Net is capable of focusing on the main subject in the image. By contrast, StrongBase has performance issues when other people appear in the front/background, dispersing its attention to other subjects in the image. This translates into a more scatter focus on the main subject, contributing to incorrect gender classification. In the image of the rightmost column, YY-Net complements facial region information with hip one, contrasting with the StrongBase, where the facial region was the main (sole) focus. The approach of YY-Net is justifiable given the background ``noise" of other people, namely an arm near the facial region, and the androgenic qualities of the subject.

\begin{figure*}[!tb]
\centering
\includegraphics[width=\textwidth]{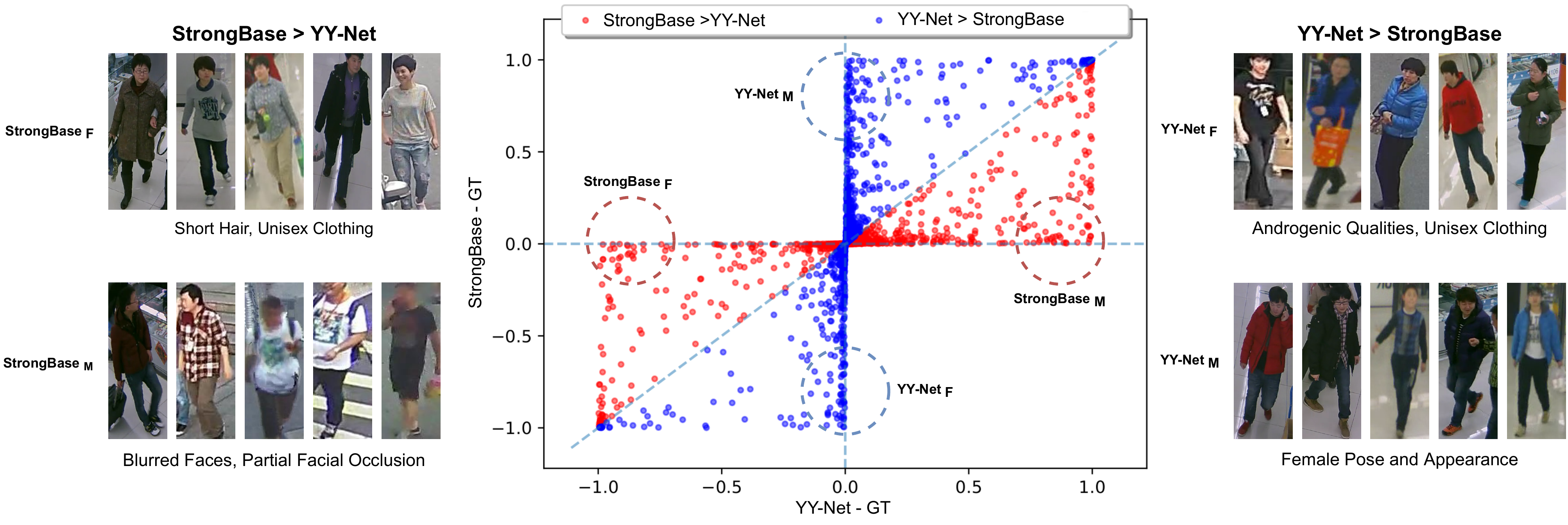}
\caption{Scatter plot of the difference between YY-Net and StrongBase predictions, and ground truth (GT) labels. Points in blue represent cases where YY-Net outperforms StrongBase, and points in red denote the opposite.  Four different regions are highlighted in the plot, displaying the cases where a model outperforms the other the most, for male (M) and female (F) recognition. Representative images, from the four regions, are presented sideways of the plot, accompanied by descriptive labels.}
\label{fig:front_fail}
\end{figure*}

To further compare YY-Net and StrongBase gender classification performance, we process the output of each model and assess the difference between each prediction and the ground truth label. We perform this comparison in a set of RAP and PA-100K images, where label 0 refers to male and 1 to female, and create a scatter plot, displayed in Fig.~\ref{fig:front_fail}. We focus our analysis on four different regions, representing cases where a model outperformed the other by the largest margins. Furthermore, we display representative images of each region, which will be the basis of our analysis. 

In the cases where YY-Net outperforms StrongBase, our model efficiently distinguishes females and males with androgenic/unisex appearance. For both genders, the pose, clothing, and face appearance misguides StrongBase in its classification, with YY-Net being more resilient. If we examine the cases where StrongBase was the outperforming model, we observe that shorter hair females, with unisex clothing, are situations where the performance of our model is subpar. Additionally, males with faces blurred, partial occluded, or with slightly uncooperative (misguided) poses might represent a challenge for YY-Net. This suggests that, although our model can handle subjects with androgenic/unisex qualities, these are also troublesome scenarios, particularly for females. For males, and females to a lesser extent, face blurriness might be a major contributing factor for misclassification. Given our subjacent approach of combining face with body information to classify gender, this could justify the underperformance in the stated conditions.

\subsection{Ablation Studies}

\begin{table}[!tb]
    \centering
    \footnotesize
    \renewcommand{\arraystretch}{1.05}
    \caption{Performance comparison, on PETA\textsubscript{Frontal} dataset, when gradually adding each component to the baseline model. Variants of the same key parts lie in the same group. Components in bold are the adopted approach in YY-Net.}
    \begin{tabular}{|c|c|}\hline
        \textbf{Component} & 
        \makebox[5.5em]{\textbf{mA}}
        \\\hline\hline
    	Baseline (Face Input) & 91.13 \\
    	Baseline (Body Input) & 92.49 \\
    	Face-Body Combination (FBC) & 92.65 \\
    	\textbf{Weighted FBC (WFBC)} & 93.10 \\ 
    	\hline
    	WFBC + Body & 93.26 \\
    	Attention + WFBC & 93.31 \\ 
    	\textbf{Attention + WFBC + Body} (\textbf{YY-Net}) & \textbf{93.45} \\
        \hline
        
    \end{tabular}
    \label{ablation-studies}
\end{table}

To validate the sensitivity of the obtained results with respect to the major components of YY-Net, we evaluate the influence of each one on gender recognition accuracy. 
As \textit{Baseline}, a ResNet50 network with the modifications described in Section \ref{network-architecture} and a classifier without the FAM are used. Furthermore, we analyze the influence of increment/variance of the components used for two key parts: 1) combination of face and body information; and 2) focus in body part attention mechanisms. We present our results in Table~\ref{ablation-studies}.

To obtain an initial baseline, we start by evaluating the importance of YY-Net components by assessing gender recognition using only face or full-body images. The results obtained by both baselines denote that full-body images (which also contain the facial region) promote a higher gender recognition accuracy than facial images. Nevertheless, neither facial nor body information by themselves are enough to obtain YY-Net performance.

We assess the effect of combining face and body information through the Hadamard product, denoted as \textit{Face-Body Combination} (FBC) in Table~\ref{ablation-studies}. This approach translates into higher $mA$ than the baseline with body input, corroborating the idea that both inputs are valuable for gender recognition in frontal images. However, this was only a slight improvement, suggesting that the combination of the input is inefficient. Adding the learnable fusion matrix $F$ to influence facial and body information combination, denominated as \textit{Weighted FBC} in the Table, promotes a $mA$ increase.

The evaluation of the attention approach (\textit{Attention}) and the preservation of complementary information (\textit{Weighted FBC  + Body}) is also a focus of our experiments. \textit{Attention} refers to the sequence of linear and nonlinear layers (described in Section~\ref{subsec:fam}) to modulate inter-channel dependencies. The residual approach of complementing the combined face-body information with body input contributes to higher gender recognition accuracy. This notion is linked to contextualizing face-body information, while using all the available information (full-body). In practice, this strategy intends to deviate the model from heavy facial region focus, regardless of the subject appearance. The attention method, along with the weighted face-body combination, provide an efficient focus on the critical parts of the body to classify gender, obtaining a similar performance to only using complementary information. Finally, \textit{YY-Net} combines the weighted FBC with attention, while also using complementary body information, which translates into an aggregated improvement when compared with all other evaluated components. This approach promotes a balanced equilibrium between face and body information, while focusing on the most influential body portions for gender recognition.

\section{Conclusions}
\label{sec:conclusion}

Subjects pose is known to hinder gender recognition in wild conditions, influencing the focus of models on different subject features. In this context, frontal images typically bias the focus towards the facial region, while the body silhouette is the main focus in non-frontal data.  Given that the state-of-the-art usually evaluates gender in good quality face datasets, we present frontal and \textit{wild} face versions of well-known PAR datasets, created through pose data processing. These subsets supply the learning data to our proposed model, YY-Net, designed to effectively complement face and body information, making it suitable for gender classification \emph{in-the-wild}. In our proposal, the key is to use a learnable fusion matrix and channel-attention sub-network, which enables a dynamic focus on the most relevant body regions relative to specific image/subject features, achieving state-of-the-art results in PAR datasets. 

The announced versions of the \textit{wild} face datasets differ from the existing ones, presenting a more challenging set to classify soft biometrics. Furthermore, the robustness displayed by YY-Net to classify gender in wild conditions supports its usability as a basis for subsequent developments in soft biometrics analysis.

\section*{Acknowledgments}

This work is funded by FCT/MEC through national funds and co-funded by FEDER - PT2020 partnership agreement under the project UIDB/50008/2020 and research grant 2020.09847.BD. 
Also, it was supported by operation Centro-01-0145-FEDER-000019 - C4 - Centro de Compet\^{e}ncias em Cloud Computing, co-funded by the European Regional Development Fund (ERDF) through the Programa Operacional Regional do Centro (Centro 2020), in the scope of the Sistema de Apoio \`{a} Investiga\c{c}\~{a}o Cient\'{i}fica e Tecnol\'{o}gica - Programas Integrados de IC\&DT.


\end{document}